# MetroLoc: Metro Vehicle Mapping and Localization with LiDAR-Camera-Inertial Integration

Yusheng Wang, *Graduate Student Member, IEEE*, Weiwei Song, Yi Zhang, Fei Huang, Zhiyong Tu and Yidong Lou

*Abstract*—We propose an accurate and robust multi-modal sensor fusion framework, MetroLoc, towards one of the most extreme scenarios, the large-scale metro vehicle localization and mapping. MetroLoc is built atop an IMU-centric state estimator that tightly couples light detection and ranging (LiDAR), visual, and inertial information with the convenience of loosely coupled methods. The proposed framework is composed of three submodules: IMU odometry, LiDAR-inertial odometry (LIO), and Visual-inertial odometry (VIO). The IMU is treated as the primary sensor, which achieves the observations from LIO and VIO to constrain the accelerometer and gyroscope biases. Compared to previous point-only LIO methods, our approach leverages more geometry information by introducing both line and plane features into motion estimation. The VIO also utilizes the environmental structure information by employing both lines and points. Our proposed method has been extensively tested in the long-during metro environments with a maintenance vehicle. Experimental results show the system more accurate and robust than the state-of-the-art approaches with real-time performance. Besides, we develop a series of Virtual Reality (VR) applications towards efficient, economical, and interactive rail vehicle state and trackside infrastructure monitoring, which has already been deployed to an outdoor testing railroad.

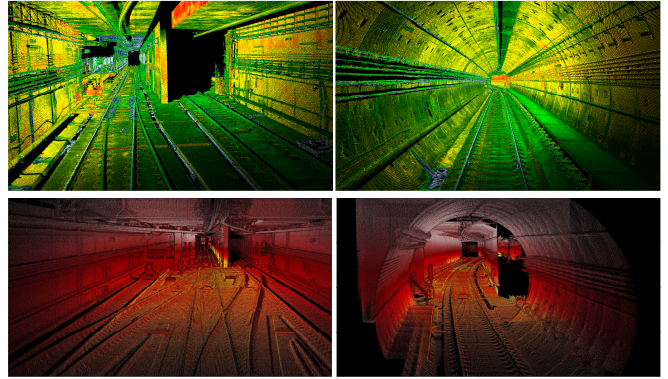

**Fig. 1.** Visual illustration of the mapping result. The above two are the ground truth generated by survey grade instruments, whereas the bottom two are from the proposed MetroLoc using a hybrid solid-state LiDAR, a supervisory camera and an IMU.

## I. INTRODUCTION

TRAIN positioning and railway monitoring is of critical importance for railroad systems since either a train localization failure or a railroad clearance intrusion might lead to fatal accidents. The train positioning strategy nowadays is dominated by trackside infrastructures like track circuits and Balise [1]. The latter approach models the train positioning as a 1D problem and divides the railway track into separate cantons, with a Balise placed at the beginning of each canton. When a train passes over one Balise, the Automatic Train Control (ATC) system knows that a train is within that canton. If another train is detected within the safe limits to this canton, the train is stopped automatically by Automatic Train Protection (ATP) system. Therefore, the train positioning accuracy is restricted by the length of a canton. Besides, the track-side infrastructure-based systems require substantial capital investment and inherent maintenance cost, where hundreds of Balises are needed in a single railway line and each Balise means thousands of dollars.

With the rapid development of sensor technologies and the worldwide standardization of railroad systems, new odometry and localization on-board sensors have found a gap in railway applications, where they can supplement the limitations of trackside ones. The precise location of the rail vehicles can be obtained through global navigation satellite system (GNSS), such as the global positioning system (GPS) and Beidou. There are numerous works on precision evaluation of GNSS for rail vehicles [2]–[4] and data fusion with IMU, odometer as well as track geometry map in [5]–[7]. These localization methods merely acquire the vehicle positioning data without extra perceptual information, which can be further utilized to operation environment monitoring and hazard forecasting. For instance, a detected crack may prevent a potential tunnel collapse and an indicated powerline failure may inhibit an accident in the future.

With the framework of estimating odometry and mapping the surroundings at the same time, simultaneously localization and mapping (SLAM) is a promising solution to this tricky conundrum. Recent advances in light detection and ranging (LiDAR) hardware have cultivated research into LiDAR-inertial fusion [8]–[10]. The accurate range measurement, invulnerability to illumination variations and long detection range of LiDARs make them suitable of navigation, localization, and mapping tasks. And they have been applied to supplement the IMU-odometer odometry in [11], [12].

*Manuscript submitted Nov 1, 2021. This work was supported by the Joint Foundation for Ministry of Education of China under Grant 6141A0211907 (*Corresponding author:* Weiwei Song).

Yusheng Wang, Weiwei Song, Zhiyong Tu and Yidong Lou are with the GNSS Research Center, Wuhan University, 129 Luoyu Road, Wuhan 430079, China (email: yushengwhu@whu.edu.cn; sww@whu.edu.cn; 2012301650022@whu.edu.cn; ydlou@whu.edu.cn).

Yi Zhang and Fei Huang are with the School of Geodesy and Geomatics, Wuhan University, 129 Luoyu Road, Wuhan 430079, China (email: yzhang@sgg.whu.edu.cn; feihuang28@whu.edu.cn).

However, in environments with degenerate geometries, such as the long metro tunnels with repetitive structures, LiDAR-only approaches may fail as frame-to-frame correspondences are estimated using highly consistent scans. Since the IMU mechanization alone cannot provide reliable pose estimates for more than a few minutes, the system failure is often nonreversible. To cope with these challenging situations, integration with extra sensors are required, and the LiDAR-visual-inertial fusion has already been successfully deployed in autonomous cave and mine exploration [13]–[15].

We are motivated to tackle the metro vehicle localization and mapping problem by LiDAR-visual-inertial fusion. There are many similarities between underground mine and metro tunnels, such as low illumination, textureless and self-repetitive. But the metro vehicle mapping and localization is a much more demanding task over tunnel exploration for three reasons:

*High speed*: The unmanned aerospace vehicles (UAV), unmanned ground vehicles (UGV) and legged robots used for fast deployment all run with low velocities. On the other hand, the metro vehicles have a regular speed of 35 km/h to 45 km/h, which raises the awareness of computation efficiency and velocity-related optimizations, such as distortion removal.

*Long journey*: Many subterrain datasets only have hundreds of meters length coverage, whereas, the distance between two metro stations is already more than one kilometer. The accumulated odometry errors will be incredibly exaggerated without proper compensation methods. Besides, metro scenes are more structured than the caves, with more than 85% column-like and repetitive tunnels.

*No loops*: Although the metro vehicles run in a circle, there are no loops for a one-way operated transportation system, thus no loop can be detected and optimized at the back-end.

Addressing the problems mentioned above, we present MetroLoc, a specific study on LiDAR-visual-inertial fusion for metro vehicle localization and mapping in this paper. The main contributions of our work can be summarized as follows:

- We propose a real-time and lightweight simultaneous localization and mapping scheme for large-scale map building in perceptually-challenging metro tunnels.
- We propose an IMU-centric LiDAR-visual-inertial fusion pipeline that tightly integrates individual measurements with constructed factor graphs.
- We propose an efficient method to extract and track LiDAR features, which significantly improve the accuracy and robustness in structured areas.
- Our proposed method is extensively evaluated in metro environments, and we also implement the approach into applications towards future railroad monitoring system.

## II. RELATED WORK

Prior works on multi-modal sensor fusion has gained great popularity using combinations of LiDAR, camera and IMU and can be classified into either loosely or tightly coupled methods. The former scheme processes the measurements from individual sensors separately, and they are preferred more for their extendibility and low computation consumption. In contrast, tightly coupled methods jointly optimize sensor measurements to obtain odometry estimation, with favorable accuracy and robustness.

### A. Loosely Coupled LiDAR-Visual-Inertial Odometry

One of the early works of LiDAR-visual-inertial, V-LOAM, is proposed in [16], which leverages the Visual-inertial odometry as the motion model for LiDAR scan matching. Since this scheme only performs frame-to-frame motion estimation, the global consistency is not guaranteed. To cope with this problem, Wang *et al* propose a direct Visual-LiDAR fusion scheme DV-loam [17]. The coarse states are estimated using a two-stage direct visual odometry module, and are further refined by the LiDAR mapping module, finally a Teaser-based [18] loop detector is utilized for correcting accumulated drifts. The robustness of the loosely coupled system can be further increased through incorporating additional constraints, such as thermal-inertial prior for smoky scenarios [19], incremental odometer [20] or legged odometry [21] for autonomous exploring robots.

### B. Tightly Coupled LiDAR-Visual-Inertial Odometry

In many of the recent works [22]–[27], tight integration of multi-modal sensing capabilities are explored for degeneracy avoidance and robustness enhancement. The filter-based approaches employs a Kalman filter for joint state estimation, such as the error state Kalman filter (ESKF) utilized in [23] and the multi-state constraint Kalman filter (MSCKF) applied in [26]. And the authors of the latter work refine a novel plane feature tracking across multiple LiDAR scans within a sliding-window, making the pose estimation process more efficient and robust. The filter-based methods are usually less extendible to other sensors and may be vulnerable to potential sensor failures. In contrast, the optimization-based approaches have proved advantageous for their expandability, where each sensor input can be encapsulated as a factor in the constructed graph.

Shan *et al* proposes LVI-SAM in [24], where the LiDAR-inertial and Visual-inertial subsystems can run jointly in feature-rich scenarios, or independently with detected failures in one of them. However, the "point to line" and "point-to-plane" based LiDAR odometry factor cost functions are not robust to feature-poor environments, and may generate meaningless result in metro or cave scenarios. To handle these situations, the geometry of 3D primitives are employed with the line and plane landmarks extracted in [22]. The robustness and accuracy of this system has been proved with two small-scale DARPA SubT datasets (one 167 m long, the other is 490 m long). In addition, the performance of Visual-inertial subsystem in structured man-made environments can also be improved with the detected line features in [28].

From the discussion of the literature, we can see that the LiDAR-visual-inertial integrated pose estimation and mapping has not been well solved and evaluated in large-scale datasets. And this paper aims to achieve real-time, low-drift and robust odometry and mapping for large-scale metro environments. To the best of the authors' knowledge, our work is the first SLAM-based metro vehicle localization and mapping system.

## III. SYSTEM OVERVIEW

The overview of our system is shown in Fig. 2, which is composed of three subsystems: IMU odometry, LiDAR-inertial odometry (LIO), and Visual-inertial odometry (VIO). Our system follows the idea of [25], where IMU is viewed as the primary sensor as long as the bias can be well-constrained by other sensors. The constrained IMU odometry provides the prediction to the VIO and LIO. The LIO and VIO submodule extracts features from raw scans and images, which are used for state estimation. Both the two modules leverage the factor graph optimization to refine the poses. Finally, the IMU odometry submodule achieves these observations to constrain the accelerometer bias and gyroscope.

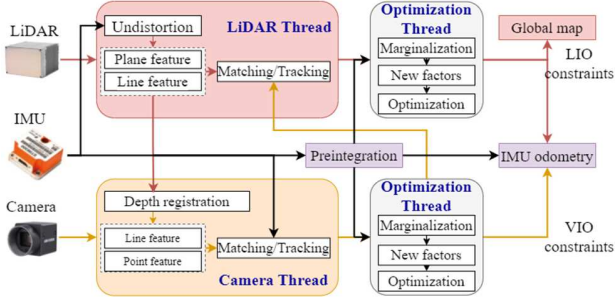

**Fig. 2.** Overview of the MetroLoc algorithm.

The notations used throughout this paper is shown in TABLE I. In addition, we define $(\cdot)^B_W$ as the transformation from world frame to the IMU frame. And the $k$-th vehicle state vector $x_k$ can be written as:

$$x_k = [\mathbf{p}^W_{B_k}, \mathbf{v}^W_{B_k}, \mathbf{q}^W_{B_k}, \mathbf{b}_a, \mathbf{b}_g] \quad (1)$$

where $\mathbf{p}^W_{B_k} \in \mathbb{R}^3$, $\mathbf{v}^W_{B_k} \in \mathbb{R}^3$, and $\mathbf{q}^W_{B_k} \in SO(3)$ are the position, linear velocity, and orientation vector. The last two elements are the IMU gyroscope and accelerometer biases.

TABLE I
NOTATIONS THROUGHOUT THE PAPER

| Notations | Explanations |
|---|---|
| | Coordinates |
| $(\cdot)^W$ | The coordinate of vector $(\cdot)$ in global frame. |
| $(\cdot)^B$ | The coordinate of vector $(\cdot)$ in IMU frame. |
| $(\cdot)^L$ | The coordinate of vector $(\cdot)$ in LiDAR frame. |
| | Expression |
| $(\hat{\cdot})$ | Noisy measurement or estimation of vector $(\cdot)$. |
| $\otimes$ | Multiplication between two quaternions. |
| $(\cdot)_{2:4}$ | Taking out the last three elements from a quaternion vector $(\cdot)$. |
| $x$ | The full state vector. |
| $e^{(\cdot)}$ | The residual of $(\cdot)$. |
| $\delta(\cdot)$ | The estimated error of $(\cdot)$. |
| $\mathbf{R}, \mathbf{q}$ | Two forms of rotation expression, $\mathbf{R} \in SO(3)$ is the rotation vector, $\mathbf{q}$ represents quaternions. |

## IV. METHODOLOGY

We use a factor graph to model this maximum a posterior (MAP) problem, and we adopt five types of factors for graph construction as shown in Fig. 3, namely: (1) IMU preintegration factors; (2) IMU odometry factors; (3) LIO factors; (4) VIO factors; (5) prior factors.

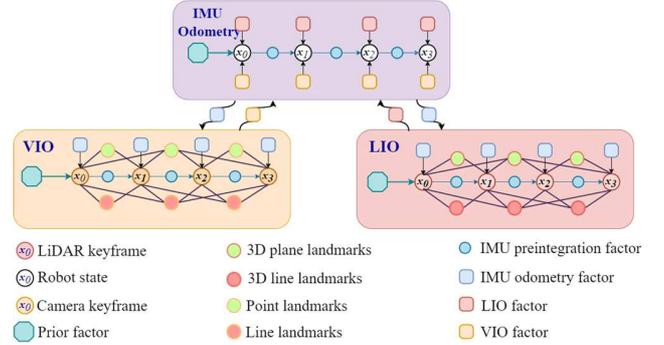

**Fig. 3.** Overview of the constructed factor graph.

### A. IMU Odometry Factors

Following [25], we add the accelerometer and gyroscope bias $\mathbf{b}_a$ and $\mathbf{b}_g$ into the estimated state of IMU odometry. The edge between two consecutive state nodes $i$ and $j$ can be obtained from the relative motion measurement. And the residual of preintegrated IMU measurements $\begin{bmatrix} \hat{\boldsymbol{\alpha}}^{B_j}_{B_i} & \hat{\boldsymbol{\beta}}^{B_j}_{B_i} & \hat{\boldsymbol{\gamma}}^{B_j}_{B_i} \end{bmatrix}$ can be formulated by:

$$e^{IMU}_{ij} = \begin{bmatrix} \delta\hat{\boldsymbol{\alpha}}^{B_j}_{B_i} & \delta\hat{\boldsymbol{\beta}}^{B_j}_{B_i} & \delta\hat{\boldsymbol{\theta}}^{B_j}_{B_i} & \delta\mathbf{b}_a & \delta\mathbf{b}_g \end{bmatrix}^T$$

$$= \begin{bmatrix} \mathbf{R}^{B_i}_W\left(\mathbf{p}^W_{B_j} - \mathbf{p}^W_{B_i} + \frac{1}{2}\boldsymbol{g}^W\Delta t^2 - \mathbf{v}^W_{B_i}\Delta t\right) - \hat{\boldsymbol{\alpha}}^{B_i}_{B_j} \\ \mathbf{R}^{B_i}_W\left(\mathbf{v}^W_{B_j} + \boldsymbol{g}^W\Delta t - \mathbf{v}^W_{B_i}\right) - \hat{\boldsymbol{\beta}}^{B_i}_{B_j} \\ 2\left[\left(\mathbf{q}^W_{B_i}\right)^{-1} \otimes \left(\mathbf{q}^W_{B_j}\right) \otimes \left(\hat{\boldsymbol{\gamma}}^{B_i}_{B_j}\right)^{-1}\right]_{2:4} \\ \mathbf{b}_{a_j} - \mathbf{b}_{a_i} \\ \mathbf{b}_{g_j} - \mathbf{b}_{g_i} \end{bmatrix} \quad (2)$$

where $\boldsymbol{g}^W = [0,0,g]^T$ is the gravity factor in the world frame and $\Delta t$ is the time sweep between $i$ and $j$. Since the VIO and LIO are both used to constrain IMU preintegration measurements, we can obtain $\mathbf{p}^W_{B_i}$, $\mathbf{v}^W_{B_i}$ and $\mathbf{q}^W_{B_i}$ from them. Besides, the bias errors are jointly optimized in the graph. With the IMU residuals $e^{LIO}_{ij}$ and $e^{VIO}_{ij}$ calculated by LIO and VIO, the IMU odometry optimization problem can be defined by:

$$E = \sum \left(e^{LIO}_{ij}\right)^T \mathbf{W}^{-1}_{ij} e^{LIO}_{ij} + \sum \left(e^{VIO}_{ij}\right)^T \mathbf{W}^{-1}_{ij} e^{VIO}_{ij} + E_m \quad (3)$$

where $\mathbf{W}_{ij}$ is the covariance matrix and $E_m$ is the marginalized prior. Since both the pose estimation of LIO and VIO are not globally referenced, we only utilize the relative state estimation as local constraints to correct the bias of IMU preintegration.

## B. LIO Factors

As illustrated in Fig. 4, we utilize a hybrid solid state LiDAR Innovusion Jaguar Prime[1] throughout our experiments. The point cloud within the 65° × 40° field of view (FoV) has a similar density coverage with that of a 300-line rotating LiDAR. The general scan matching method is not suitable for this scenario for lack of computation efficiency. Besides, employing all the feature points for state estimation is not robust towards degeneracy problems.

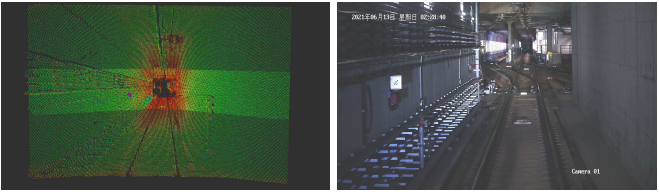

**Fig. 4.** Visual illustration of the scan pattern of Innovusion Jaguar Prime. The left inset is the raw point cloud of a single frame and the right one indicates the corresponding image.

We notice that the metro tunnel is a highly structured scenario with many line features, for instance, rail tracks, power lines, and pipelines. Such line features have proved to be a promising approach towards man made structured scenarios [28], [29]. Since the rail tracks are the most distinctive line structures in the point cloud, they are first extracted based on the geometric patterns.

We first determine the track bed area using the LiDAR sensor mounting height. The track bed is composed of rail sleepers, rail tracks, and track-side sensors, where the rail tracks are the highest infrastructures. With the assumption of the LiDAR is centered between two rail tracks, we can set two candidate areas around the left and right rail tracks and search the points above a certain threshold over the track bed. Two straight lines can then be fixed using random sample consensus (RANSAC) [30] method. Finally, we exploit the idea of region growing [31] for further refinement. As a prevailing segmentation algorithm, region growing examines neighboring points of initial seed area and decides whether to add the point to the seed region or not. We set the initial seed area within the distance of 3 m ahead of the LiDAR, and the distance threshold of the search region to the fitted line is set to 0.07 m, which is the width of the track head. Note that we only extract the current two tracks where the metro vehicle is on, and a maximum length of 15 m tracks are selected for each frame.

The other line features are extracted following the depth-continuous edge extraction method proposed in [32]. The point cloud is divided into small voxels for each input frame, and planes are fitted repeatedly in the voxels. Then the depth-continuous edges can be detected at the plane intersections. We sample multiple points on each extracted line including the two tracks, and retain the lines that satisfy the infinite straight line model [33]. The extracted lines are shown in Fig. 5 thanks to the high point cloud density coverage.

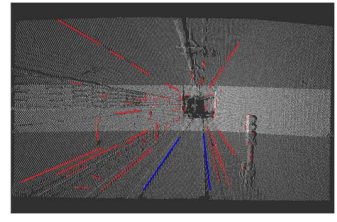

**Fig. 5.** Visual illustration of the extracted line features, blue indicates the rail tracks and red describes other lines.

The infinite straight lines can be parameterized by a rotation matrix $\mathbf{R}$ and two scalars $a, b \in \mathbb{R}$. Let the error operator $\ominus$ between two lines $\mathbf{L}_i, \mathbf{L}_j$ be defined as:

$$\mathbf{L}_i \ominus \mathbf{L}_j = \left( \begin{bmatrix} 1 & 0 \\ 0 & 1 \\ 0 & 0 \end{bmatrix}^T Log(\mathbf{R}_i^T \mathbf{R}_j), a_i - a_j, b_i - b_j \right) \quad (4)$$

Then the residual between the two lines can be formulated by:

$$\mathbf{e}_{ij}^{LIO_{li \to li}} = \left( \mathbf{T}_{L_i}^{L_j} \mathbf{L}_i \right) \ominus \mathbf{L}_j \quad (5)$$

where $\mathbf{T}_{L_i}^{L_j} = (\mathbf{R}_i^j, \mathbf{p}_i^j)$ describes the transformation between the two lines. And we use the derivatives of equation (5) in the optimization process with symmetric difference method.

Additionally, a fitted plane $\mathbf{P}$ can be parameterized by the united normal direction vector $\mathbf{n}_P$ and a distance scalar $d_P$, such that $\mathbf{P} = [\mathbf{n}_P^T, d_P]^T$. And the measurement residual from a point to a plane can be expressed as:

$$\mathbf{e}^{LIO_{po \to pl}} = \mathbf{T}_{P_i}^{p_j} \mathbf{P}_i \quad (6)$$

Suppose we have $N_{pP}$ sets of point-to-plane correspondences and $N_p$ points, the plane-to-plane residual can be formulated by:

$$\mathbf{e}^{LIO_{pl \to pl}} = min \sum_{i=1}^{N_{pP}} \sum_{j=1}^{N_p} \mathbf{e}_{ij}^{LIO_{po \to pl}} \quad (7)$$

We use the second-order derivatives of point-to-plane cost for efficient optimization as illustrated in [34]. In addition, we use the IMU forward propagation to predict the location of the previous lines and planes in the current scan. Two lines are considered a match if their directions and center distances divergences are lower than 10° and 0.5 m. Similarly, two planes are regarded as a match when difference between their normal and the distance scalar are smaller than 5° and 0.25 m.

Given equation (2), (5) and (7), the minimization problem of LIO can be expressed as follows:

$$min \left\{ \sum_{i \in N_L} \mathbf{e}_i^{LIO_{li \to li}} + \sum_{i \in N_P} \mathbf{e}_i^{LIO_{pl \to pl}} + \sum_{i \in N_I} \mathbf{e}_i^{IMU} + \mathbf{e}_m^{LIO} + \mathbf{e}_{imuodom}^{prior} \right\} \quad (8)$$

where $N_L$, $N_P$, and $N_I$ are the number of line-to-line, plane-to-plane correspondences, and IMU preintegration factors. $\mathbf{e}_m^{LIO}$ is the marginalization factors and $\mathbf{e}_{imuodom}^{prior}$ is the predicted pose constraints from IMU odometry. The optimization problem can be efficiently solved by Levenberg-Marquardt algorithm [35].

*C. VIO Factors*

Our VIO follows the pipeline of Vins-mono [36] as shown in Fig. 3. The point features are detected by [37], tracked by KLT sparse optical flow [38], and refined by RANSAC. For the line features, we employ the LSD [39] for line segment extraction and PL-VINS [40] for line feature fused state estimation as shown in Fig. 6(a). In addition, we register LiDAR frames to the camera frames, and project the 3D distortion-free point cloud to the 2D image as illustrated in Fig. 6(b). Then we can use the depth information for scale correction and joint graph optimization following [28]. And the minimization problem of VIO can be expressed as follows:

$$\min\left\{\sum_{i\in N_{li}} e_i^{VIO_{li\to li}} + \sum_{i\in N_{po}} e_i^{VIO_{po\to po}} + \sum_{i\in N_I} e_i^{IMU} + e_m^{VIO} + e_{imuodom}^{prior}\right\} \quad (9)$$

where $e_i^{VIO_{li\to li}}$, $e_i^{VIO_{po\to po}}$, and $e_m^{VIO}$ are the relative line and point reprojection error, and marginalization factors. $N_{li}$ and $N_{po}$ represent the amount of line and point features tracked by other frames.

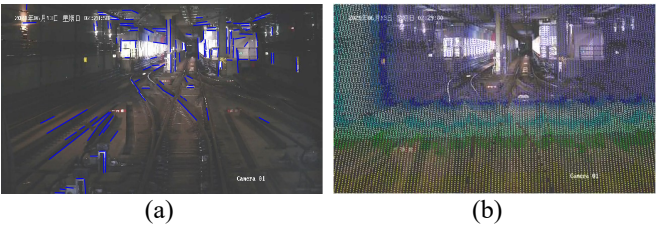

(a)            (b)

**Fig. 6.** Visual illustration of the extracted line feature in (a) and the depth frame in (b), the color is coded by the depth value.

## V. EXPERIMENTS

*A. Hardware Setup and Ground Truth Description*

We select a metro maintenance vehicle as shown in Fig. 7, which includes a hybrid solid-state LiDAR Innovusion Jaguar Prime, a tactical grade IMU Sensonor stim 300, and a Hikvision supervisory camera. All the three sensors are hardware synchronized with an external atomic clock. The dataset is processed by a high-performance onboard computer, with i9-10980HK CPU, 64 GB RAM. Besides, our algorithms are implemented in C++ and executed in Ubuntu Linux using the ROS [41]. Since the output of the Hikvision camera is in the video format, we first merge the video with the recorded dataset using the synchronized timestamp. The nonlinear optimization and factor graph optimization problem is solved using Ceres Solver and GTSAM [42], respectively. Our proposed system can reach real-time performance for all the captured datasets.

The odometry and map ground truth is captured by two survey grade 3D laser scanners, a near navigation grade IMU containing fiber optic gyros, and processed by professional post-processing software. The initial global position is set by Real Time Kinematic (RTK) at the starting point outdoors.

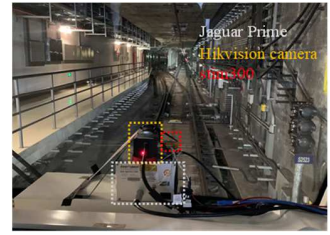

**Fig. 7.** The hardware setup of our system.

We conduct extensive experiments utilizing the maintenance vehicle platform, and two novel LiDAR-visual-inertial fusion algorithms are selected for comparison: LVI-SAM [24] and R2LIVE [23]. In addition, LOAM [10], Lio-sam [9], and VINS-Mono [36] are also selected for ablation study here. The LIO and VIO part of our system is denoted as MetroLIO and MetroVIO, respectively.

*B. Experiment-1: Feature-Rich Crossover and Station*

Depicted in Fig. 8(a) and Fig. 8(b), the metro crossover and station are feature-rich areas with multiple rail tracks, guardrails, pipelines, and platforms. Besides, the visual sequences are not degraded because of the head light and station lighting.

In this experiment, we manually drive the metro maintenance vehicle for 420 m with a relative low velocity. TABLE II shows the evaluation results comparing to different approaches. And we can infer that the short-during scenario seems to be effortless for state-of-the-art SLAM algorithms.

*C. Experiment-2: Structured Tunnel*

In this experiment, we aim to evaluate the robustness of our algorithm under a 750 m metro tunnel scenario, where the only observable features are repetitive and structured man-made infrastructures as shown in Fig. 9(a) and Fig. 9(b). Since the tunnel walls are almost flat and no extra dynamic objects can be employed for feature matching, the metro tunnel is one of the most challenging scenarios for SLAM.

The state estimation errors of different methods are listed in TABLE II. Due to the highly regulated and constrained motion on the rail tracks, the IMU biases cannot be initialized and estimated correctly with inadequate axis excitement. This leads to serious scale drift and result in significant pose estimation errors in VINS-Mono. Besides, the wrong feature matches from repetitive pattern of the surround also exacerbate the deviations.

The feature-based LiDAR SLAM also suffers from feature tracking problems. With only frame-to-frame matching, LOAM fails several times in the tunnel, and the LiDAR odometry even moves backwards without accurate pose estimation. Since Lio-sam relies heavily on the LiDAR odometry to further constrain the pre-integrated IMU states, it also generates unfavorable results. On the contrary, both the MetroLIO and MetroVIO achieves less drift state estimation employing plane-line and point-line feature tracking.

TABLE II
ACCURACY EVALUATION OF EXPERIMENT-1 AND EXPERIMENT-2

| Sequence | Absolute Trajectory Error (ATE) – Translational [m] / Rotational [°] | | | | | | | |
|---|---|---|---|---|---|---|---|---|
| | LVI-SAM | R2LIVE | LOAM | Lio-sam | VINS-Mono | MetroLIO | MetroVIO | MetroLoc |
| Experiment-1 | 0.49 / 2.56 | **0.32 / 1.80** | 1.45 / 3.68 | 0.53 / 1.75 | 4.23 / 6.92 | 0.58 / **1.64** | 1.74 / 4.98 | 0.63 / 2.72 |
| Experiment-2 | 22.79 / 45.79 | 8.43 / 9.87 | 48.75 / 26.23 | 38.92 / 22.00 | 80.94 / 55.67 | 4.75 / 6.41 | 7.62 / 8.07 | **1.92 / 3.58** |

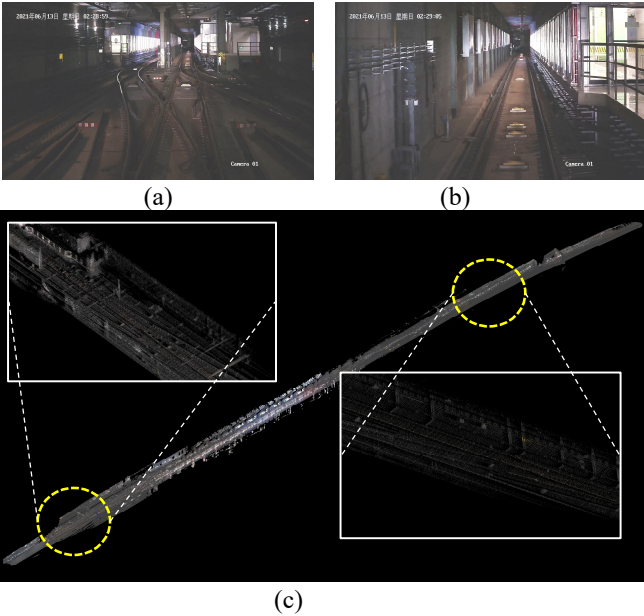

**Fig. 8.** Visual illustration of the crossover and station. (a) and (b) denotes the images from a crossover and a metro station, respectively. (c) is the MetroLoc mapping result aligned with RGB color.

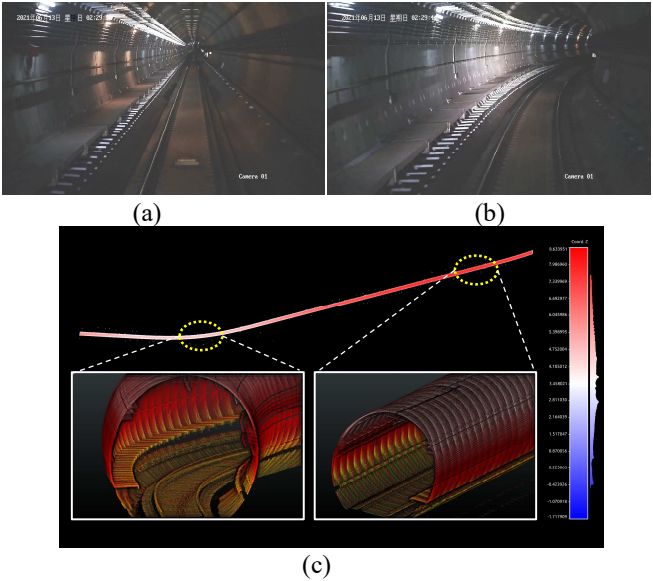

**Fig. 9.** Visual illustration of the structured tunnel. (a) and (b) are images from the supervisory camera. (c) denotes the vertical view of the mapping result from MetroLoc.

[2] https://github.com/MichaelGrupp/evo

*D. Experiment-3: Long-During Mapping and Odometry*

In this experiment, we aim to show that MetroLoc is accurate and robust enough to reconstruct a long-during map in metro environments. The overall length is around 5.2 km with 4 stations along the path. We plot the real-time reconstructed 3D maps as well as the ground truth in Fig. 10, where both the horizontal headings and height variations are visually well matched. Note that there are many outlier points around tunnel structures in the reconstructed map, we believe they are mainly caused by the LiDAR range measurement noises. In addition, we adopt the EVO[2] package to compare each trajectory against ground truth as shown in Fig. 11. We can infer that the monocular vision-only methods have the worst performance in this situation, as explained above. Our MetroLoc outperforms the selected methods with the lowest ATE of 5.62 m.

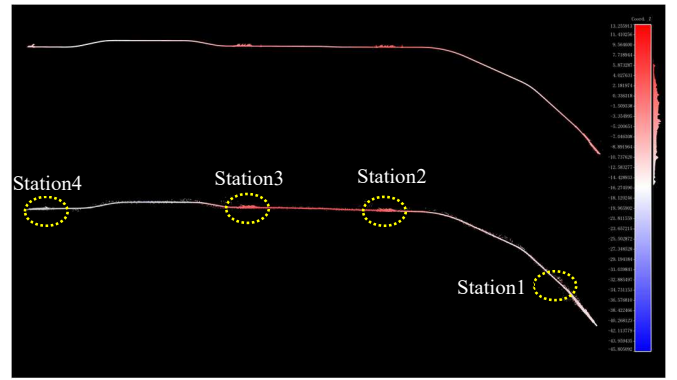

**Fig. 10.** Visual comparison of the mapping result from ground truth (above) and MetroLoc (bottom), the color indicates height variations. The blue areas are filled by the outlier points.

TABLE III
THE AVERAGE RUNNING TIME OF SUBMODULES [MS]

| Data | MetroLIO | MetroVIO | IMUODOM |
|---|---|---|---|
| Experiment-1 | 37 | 21 | 0.5 |
| Experiment-2 | 32 | 19 | 0.5 |

*E. Runtime Analysis*

Thanks to the multi-threaded computation, all the subsystems can run in parallel. The average time consumption of each submodule is listed in TABLE III, illustrating that MetroLoc can achieve real-time performance for the onboard computer.

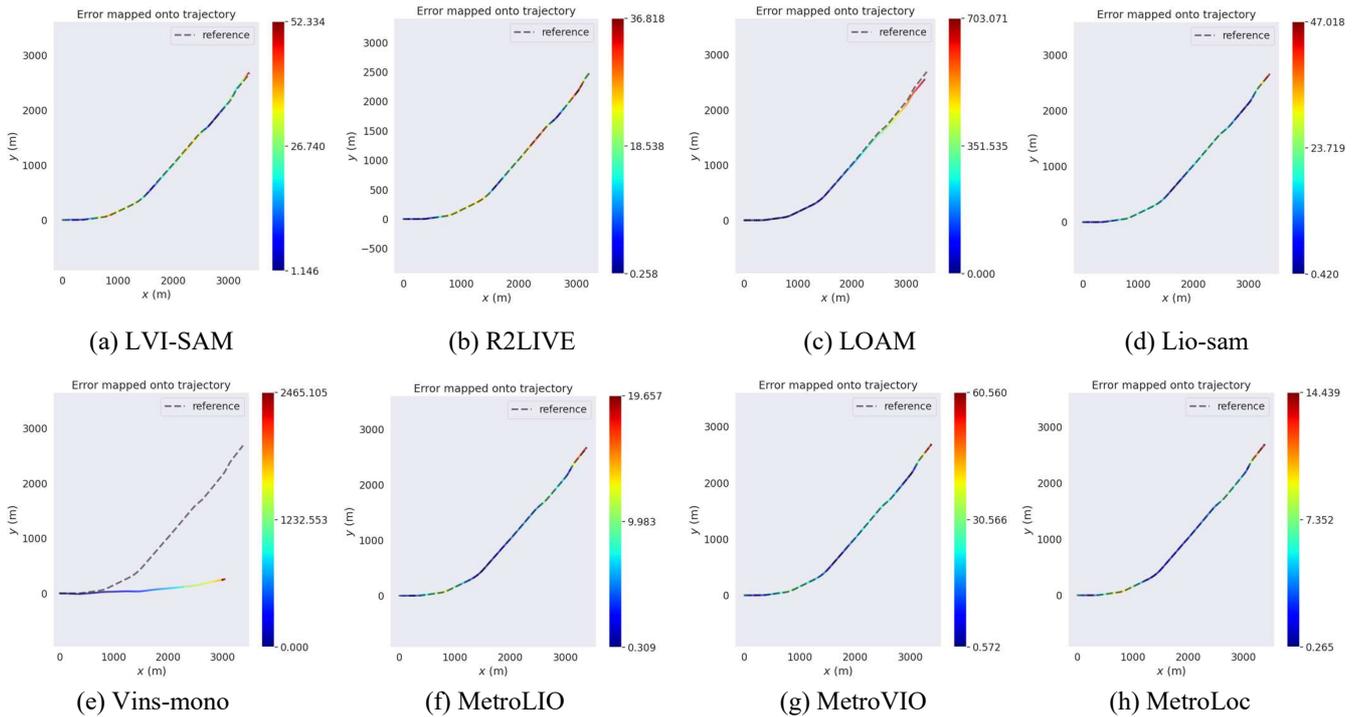

**Fig. 11.** Trajectory evaluations of different methods.

*F. Towards Next-generation Railway Monitoring*

The current railway and metro rail maintenance system is still dominated by human force. For the metro system, the workers always drive the maintenance vehicle at midnight to manually check the safety of power line, pipeline, as well as trackside infrastructures. We hereby propose a prototype design of a semi-auto monitoring system using the reconstructed map from MetroLoc. The maps are exported to a game engine Unity[3], which can be easily visualized through Virtual Reality (VR) glasses, for instance, HTC VIVE Pro. Since the 3D points all have pose information, the operator can easily locate the potential districts. Besides, the abnormal powerlines can be indicated with integrated infrared cameras. We also develop various tools in VR, such as direct range measuring, region picking, and view sharing. In addition, we can use the AirSim[4] and the pose estimation from MetroLoc to visualize the real-time train location information as illustrated in Fig. 12.

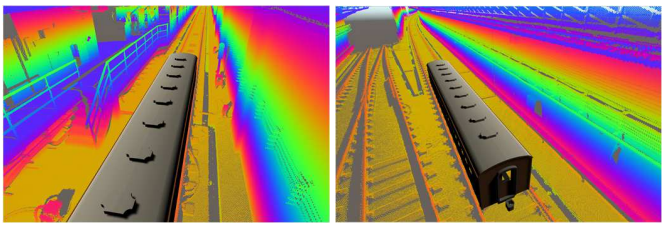

**Fig. 12.** Visual illustration of the train pose monitoring system, the color is coded by height variations.

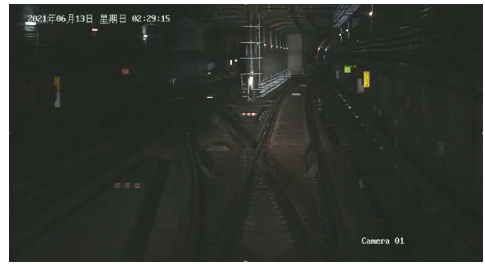

**Fig. 13.** Visual illustration of the hectometer post (recognized and marked) and station indicator signs (yellow) on the wall.

## VI. CONCLUSION

In this paper, we propose a robust and versatile simultaneous localization and mapping approach for metro vehicles. Our method tightly integrates measurements from LiDAR, camera, and IMU with an IMU-centric state estimator. To cope with the highly repeated man-made structures, we propose to extract and track geometric features. The proposed method has been validated in extreme metro environments. And the results show that our framework is accurate and robust towards long-during motion estimation and mapping problems. We hope that our experimental work and extensive evaluation could inspire follow-up works to explore more structural or situational awareness information in highly-repetitive scenarios, such as tags or signs on the wall as shown in Fig. 13. Besides, we wish the community pay more attention to railroad applications, especially for facility and environment monitoring. A small advance towards autonomous system will save tremendous amount of manpower for construction and maintenance.

---

[3] https://unity.com/

[4] https://microsoft.github.io/AirSim/

ACKNOWLEDGMENT

We would like to thanks colleagues from CRRC Zhuzhou Locomotive CO., LTD for providing testing platform and gathering experimental data.